\documentclass[conference]{IEEEtran}
\IEEEoverridecommandlockouts
\usepackage{algorithm, algorithmic}
\usepackage{soul}
\usepackage{cite}
\usepackage{amsmath,amssymb,amsfonts}
\usepackage{algorithmic}
\usepackage{graphicx}
\usepackage{textcomp}
\usepackage{xcolor}
\def\BibTeX{{\rm B\kern-.05em{\sc i\kern-.025em b}\kern-.08em
    T\kern-.1667em\lower.7ex\hbox{E}\kern-.125emX}}

\begin{document}

\title{Between-Domain Instance Transition Via the Process of Gibbs Sampling in RBM
}

\author{\IEEEauthorblockN{Hossein Shahabadi Farahani}
\IEEEauthorblockA{\textit{APAC Research Group} \\
\textit{Dept. Mechatronics} \\
\textit{K.N.Toosi University of Technology}\\
Tehran, Iran \\
shahabadi.f@email.kntu.ac.ir}
\and
\IEEEauthorblockN{ Alireza Fatehi}
\IEEEauthorblockA{\textit{APAC Research Group} \\
\textit{Dept. of System And Control} \\
\textit{K.N.Toosi University of Technology}\\
Tehran, Iran \\
fatehi@kntu.ac.ir}
\and
\IEEEauthorblockN{Mahdi Aliyari Sh}
\IEEEauthorblockA{\textit{APAC Research Group} \\
\textit{Dept. Mechatronics} \\
\textit{K.N.Toosi University of Technology}\\
Tehran, Iran \\
aliyari@kntu.ac.ir}

}

\maketitle

\begin{abstract}

In this paper, we present a new idea for Transfer Learning (TL) based on Gibbs Sampling.
Gibbs sampling is an algorithm in which instances are likely to transfer to a new state with higher possibility with respect to a probability distribution. We find that such an algorithm can be employed to transfer instances between domains.
Restricted Boltzmann Machine (RBM) is an energy based model that is very feasible for being trained to represent a data distribution and also for performing Gibbs sampling. We used RBM to capture data distribution of the source domain and use it in order to cast target instances into new data with a distribution similar to the distribution of source data. Using dataset that are commonly used for evaluation of TL methods, we show that our method can successfully enhance target classification by a considerable ratio.
Additionally, the proposed method has the advantage over common DA methods that it needs no target data during the process of training of models.

\end{abstract}


\section{Introduction}
Human being is able to use the knowledge collected in learning a task to learn other new tasks more efficiently. In this regard, the idea of TL is to imitate such a ability in machine learning problems \cite{pan2009survey}. With the perspective of machine learning, the discrepancy of tasks relies on the difference of the data distribution. Therefore, TL algorithms helps to extract knowledge from a domain of data and adapt it to another domain with different data distribution \cite{weiss2016survey}. The domain from which he knowledge is collected is called source domain and the domain in which the knowledge is utilized is called target domain.

TL has raised large interest in machine learning community, since it is a common practical challenge in many machine learning application that the distribution of labeled data by which the models are trained is not the same as the distribution of data to which the models are applied. In the context of image processing, when the image background, color mode or orientation of object in images are changed the models can not classify images accurately anymore \cite{ganin2016domain}. In robotic applications, it is challenging to apply policies and controller learnt in simulation environment to real world robots due to the difference in their perception or dynamics \cite{RoboImitationPeng20, chen2018transferring}. Intelligent machine health monitoring systems also suffers from the fact that machines' behavior vary in different working conditions \cite{shao2018highly}. All the mentioned problems has been tackled to by TL algorithms. 


TL algorithms can be categorized into three different groups. The most common approaches of TL are parameter-based methods, which try to use  parameters of a model trained in the source domain, to make a relatively better model for the target domain, usually by fine-tuning the model using limited available labeled data from target \cite{donahue2014decaf}. Another group of TL methods are instance-based methods. In this methods, source samples are weighted based on their similarity to the target instances which is measured in a probabilistic sense. Then this weights are used to train a model that is more adapted to the target domain \cite{lin2013double}. Finally, the most related to the method that is proposed in this paper, are representation learning based methods which also called feature-based methods. 

In representation learning based TL techniques, the goal is to find a mapping that minimizes the label predication risk in the source domain, while reducing a notion of distance between the source and target domain \cite{ganin2016domain}.
Tzeng et al \cite{tzeng2014deep} measured between domains distances of source and target samples via Maximum Mean Discrepancy (MMD). This criterion is widely used in many other researches later \cite{lu2016deep,long2017deep}. In these methods, a neural networks is trained to minimize label prediction cost while being penalized by MMD of features extracted from the source and target data.
Ben-David et al \cite{ben2010theory} show that the generalized error of classification between the source and target samples can be interpreted as a score of divergence of domains. It is theoretically proved that assuming there exist a representation of source and target domain in which their divergence is minimized, a classifier with a low error (known as risk) for the source domain will have a better performance on target domain relatively.
In this regard, domain adversarial training methods tries to learn a representation on which the source and target data are indistinguishable by adversarial training of the feature extractor and a domain classifier \cite{ganin2016domain}. All these representation learning based TL techniques trains a feature extractor, usually an Artificial Neural Network (ANN), that maps source and target data into a mutual subspace in which their distribution are closed to each other. 

The proposed method utilizes the RBM which is an energy based model that represents a  Joint Probability Distribution (JPD) over a set of variables through its energy function. RBM is a restricted version of Boltzmann machine, however it is highly capable for representing complex data distributions\cite{salakhutdinov2009deep} and there exit many efficient algorithms for training it \cite{fischer2014training}. Besides, it can be easily employed to perform Gibbs sampling for getting sample from the represented JPD. Gibbs sampling is a process in which instances are transferred by influence of the gravity of energy minima of a energy surface. In fact, Gibbs sampling leads a current state to a new sample with higher portability with respect to a desired probability distribution.


    Despite the fact that representation learning based TL methods transfer instances into new domains using a deterministic model, In the proposed method such a mapping happens via the process of Gibbs sampling. Employing Gibbs sampling as a stochastic transition process for transferring samples among different domains is a novel idea which is presented by this paper for the first time. We suggest to train a RBM based on source data in order to use it as a generative model that transfers target samples into the source domain. Using the proposed method, we could perform pixel-level TL which has been addressed by few researches \cite{hoffman2017cycada}.
    Furthermore, in the proposed method, target samples are not involved in the training of models and they are used only in the prediction stage while DA techniques usually require huge amount of unlabeled target data during training of the models. Consequently, our assumption about the availability of data is far more restricted compere to common DA problems which means our algorithm can be employed in more difficult situations.

\section{Boltzmann Machine}
A Boltzmann machine can be defined with different perspectives. For instance, it can be defined as a stochastic Hopfield network or recurrent neural networks with the point of view of ANNs \cite{barra2012equivalence}. For the best understanding of how Boltzmann machine represents data distribution and its properties, it is needed to look at it as a probabilistic graphical models. Therefore, in this paper, we mostly look at this model as a Markov random field or undirected probabilistic graphical model. In this section, the structure of Boltzmann machine and RBM is introduced. Then it is explained that how RBM represents a joint probability distribution. Finally, the concept of Gibbs Sampling is clarified.

\subsection{General Boltzmann Machine}
\begin{figure}[htbp]
\centerline{\includegraphics[width=1.5in]{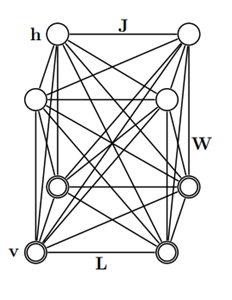}}
\caption{The architecture of general Boltzmann machine\cite{salakhutdinov2009deep}.}
\label{bm}
\end{figure}

     The general structure of Boltzmann machine is presented in the Fig. \ref{bm} \cite{salakhutdinov2009deep}. This model is consisted of numbers of units or neurons whose values are  0 or 1. As shown in the figure, these units are divided into two groups, hidden units and visible units, which is explained later. This is a  fully connected network which means  in this model each unit is connected to all other units. In fact, the structure shown in Fig. \ref{bm} is a graphical representation of a Markov random field which represents JPD of all visible and hidden variables through an energy function. More detail regarding the energy function and representation of the model is provided in the next section. 

\subsection{Restricted Boltzmann Machine}\label{sec_rbm}
\begin{figure}[htbp]
\centerline{\includegraphics[width=1.5in]{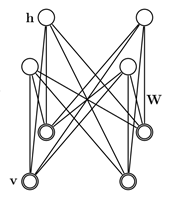}}
\caption{The architecture of restricted Boltzmann machine \cite{salakhutdinov2009deep}.}
\label{rbm}
\end{figure}

If all within connections of hidden units and all within connections of the visible units of Boltzmann machine are removed, the new appeared structure is like Fig. \ref{rbm} and is called RBM \cite{salakhutdinov2009deep}. Alternatively, RBM is a Boltzmann machine in which no two hidden units and no visible units are connected. 

According to the weights of units connection, $W$, and bias weights of visible and hidden units, $b$ and $c$, the negative energy of RBM is calculated by formula (\ref{e}) given the value visible and hidden units, $v$ and $h$:

\begin{equation}\label{e}
    \begin{aligned}
    -E\left( v,h \right)&=\underset{v}{\overset{{}}{\mathop \sum }}\,{{b}_{i}}{{v}_{i}}+\underset{h}{\overset{{}}{\mathop \sum }}\,{{c}_{i}}{{h}_{i}}+\underset{v,h}{\overset{{}}{\mathop \sum }}\,{{W}_{ij}}{{v}_{i}}{{h}_{j}}\\&={{b}^{T}}v+{{c}^{T}}h+{{v}^{T}}Wh
    \end{aligned}
\end{equation}

Briefly, the intuition behind the energy distribution is that the higher the energy of a $(v,h)$ is the lower its probability will be.

Considering energy function (\ref{e}) the JPD 
can be obtained from (\ref{jpd}):
\begin{equation}\label{jpd}
    \begin{aligned}
        P\left( v,h \right)=\frac{1}{Z}\text{exp}\left( -E\left( v,h \right) \right)\\
        Z=\underset{v}{\overset{{}}{\mathop \sum }}\,\underset{h}{\overset{{}}{\mathop \sum }}\,\text{exp}\left( -E\left( v,h \right) \right)\\
    \end{aligned}
\end{equation}

The term $Z$, which called partition function, is usually impractical to be calculated, and so, it is estimated through Annealed Importance Sampling (AIS) algorithm \cite{neal2001annealed}. 

Visible variables, $v$, are related to features or variables of the dataset to which the RBM is fitted, where the ultimate goal of the RBM is to represent their distribution. The marginal probability distribution of visible variables, $P(v)$, is calculated by (\ref{marginal}):
\begin{equation}\label{marginal}
    \begin{aligned}
        P\left( v \right)=\underset{h}{\overset{{}}{\mathop \sum }}\,P\left( v,h \right)
    \end{aligned}
\end{equation}
Actually, hidden variables are used to enrich the energy function so that the model can represent the distribution of visible variables.

 As mentioned before, it is hard to compute the value of the partition function, $Z$, in the Boltzmann machine which is required for computing exact probabilities, training of the model, inference and sampling.
 Nonetheless, when the structure of Boltzmann machine is modified into RBM, many of the practical problems about BM can be handled. As shown by (\ref{conditionally indepentdent}), given the value of hidden units, $h$, each visible variables, ${{v}_{i}}$, is conditionally independent from other visible variables. 
 Besides, these cumulative probably functions CPDs can simply be calculated by a sigmoid function like (\ref{sigmoid}).
 
\begin{equation}\label{conditionally indepentdent}
    \begin{aligned}
        p(v|h)&=\prod\limits_{i=1}^{n}{p({{v}_{i}}|h)}
    \end{aligned}
\end{equation}
Consequently, it is easy to sample from the this conditional probability.
\begin{equation}\label{sigmoid}
    \begin{aligned}
        p({{v}_{i}}=1|h)=sigmoid({{b}_{i}}+{{W}_{i:}}h)
    \end{aligned}
\end{equation}
 In a similar way, given the value of visible units, the CPD of all the hidden units can be computed using simple sigmoid functions independently.
 
 The fact that RBM is very feasible for calculating CPDs and consequently, getting sample from its units, have made it a popular model. Sampling from these conditional dependencies is a basic step in almost all of the proposed RBM learning algorithms. Furthermore, most of the applications of RBM are based on sampling. 
 
 Training of RBM is an unsupervised learning algorithm, which means only unlabeled data are used. In the training of a RBM, the goal is to adjust the model parameters, $\{W,b,c\}$, such that the energy function (\ref{e}) represents a JPD, in which training data, ${v}^{(t)}$, have the maximum likelihood. The log-likelihood function for training of RBM is as (\ref{likelihood}) \cite{fischer2014training}.
\begin{equation}\label{likelihood}
    \begin{aligned}
        \ell (\{W,b,c\}; {v}^{(t)})&=
        \sum\limits_{t=1}^{n}{\log P({{v}^{(t)}})}=
        \sum\limits_{t=1}^{n}{\log \sum\limits_{h}^{{}}{P({{v}^{(t)}},h)}}\\
        &=\sum\limits_{t=1}^{n}{\log \sum\limits_{h}{\exp \{-E({{v}^{(t)}},h)\}-n\log Z}}
    \end{aligned}
\end{equation}

 Gradient descent algorithm can be employed to maximize $\ell (\{W,b,c\}; {v}^{(t)})$. However, the partition function, $Z$, in the second term of this equation makes it impractical to compute the exact gradients. All algorithms proposed for training of RBM, contribute to finding a solution for this obstacle. Contrastive Divergence (CD) \cite{hinton2002training} and Persistent Contrastive Divergence (PCD) \cite{tieleman2008training} are most common approaches that are used for training RBM. It is important to keep in mind that in CD algorithm, the energy surface of the RBM is only modified in very close neighborhood of training instance. Therefore, the energy function of a RBM that is trained by CD does not represent the distribution of training data.

 \subsection{Gibbs Sampling}
\begin{figure}[htbp]
\centerline{\includegraphics[width=2.5in]{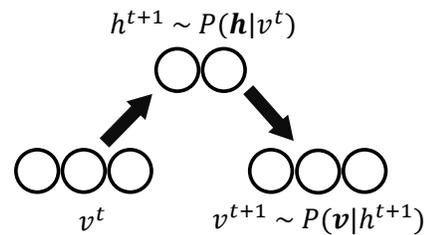}}
\caption{Transition form $v^t$ to $v^t+1$ by one step Gibbs sampling in RBM.}
\label{rbm_Gibbs sampling}
\end{figure}

Gibbs sampling is the process of a stochastic transition from an initial state, $x^{t}$, to another state, $x^{t+1}$, regarding a desired JPD, $P(X)$, through the Algorithm \ref{alg}. As mentioned before, it is feasible to compute the CPD of visible units of a RBM, $v$, given the value of hidden units, $h$ with a sigmoid function and vice versa. Therefore, based on the  Algorithm \ref{alg}, given an initial state,$(v^{t},h^{t})$, it is easy to Gibbs sample $(v^{t+1},h^{t+1})$. As shown by Fig. \ref{rbm_Gibbs sampling}, first of all, the updates of hidden units are simultaneously sampled from $P(h|v^{t})$. Afterward, updates of all visible units are sampled from $P(v|h^{t+1})$. 
Also, it is important to keep in mind that when using a RBM for Gibbs sampling, the desired probability distribution is the JPD, which is represented by the energy surface of the RBM.

 \begin{algorithm}
 \caption{One step transition by Gibbs sampling}
 \begin{algorithmic}[1]\label{alg}
 \renewcommand{\algorithmicrequire}{\textbf{Require:}}
 \renewcommand{\algorithmicensure}{\textbf{Ensure:}}
 \REQUIRE $(x^{t}_1, ..., x^{t}_n)$, $ P(X_1, ..., X_n)$
 \ENSURE  $(x^{t+1}_1, ..., x^{t+1}_n)$
  \FOR {$i = 1$ to $n$}
  \STATE Sample $x^{t+1}_i \sim P(X_i|x^{t}_{-1})$
  \ENDFOR
 \RETURN $(x^{t+1}_1, ..., x^{t+1}_n)$ 
 \end{algorithmic} 
 \end{algorithm}

Actually, Gibbs sampling is among Markov Chain Monte Carlo (MCMC) algorithms which are used to get sample from a desired probability distribution, $P(X)$ \cite{koller2009probabilistic}. The general principle in these algorithms is to do a chain of transitions, using an algorithm like Gibbs sampling, up until reaches to thermal equilibrium \cite{koller2009probabilistic}. In thermal equilibrium, the current state is a sample from the desired probability distribution. For instance, RBM can generate samples similar to its training data by using Gibbs sampling in chain \cite{salakhutdinov2009deep} \cite{keyvanrad2014brief}. In this paper, we do not go in details of the thermal equilibrium and sampling from a desired distribution. However, for understanding of how our proposed method works, it is important to have an intuition of what happens in every single step of Gibbs sampling.

When RBM is used to perform one step Gibbs sampling, $E(v^{t+1},h^{t+1})$ is probably lower than $E(v^{t},h^{t})$. It can be imagine that areas with lower energy have higher gravity that attract samples during Gibbs sampling. For example, when the current state of a RBM is near a local maximum energy, it is very likely to escape to lower energy neighborhood areas after a step of Gibbs sampling. On the contrary, when the current state is near the low energy areas, it is very likely to preserve its state and would not probably transfer to higher energy neighborhoods after Gibbs sampling. However, sampling is a stochastic phenomenon, thus, transitions might not be in alliance with the mentioned tendencies. 

             \section{Proposed Method}

 In this paper, the problem of TL is tackled to with novel perspective. In the proposed method, samples are cast into a new domain via the process of Gibbs sampling while DA methods usually align features extracted across domains using a deterministic mopping, usually an ANN\cite{ganin2016domain, tzeng2014deep, long2017deep, ghifary2016deep}.
     Furthermore, almost all DA methods require unlabeled target data for training domain adaptive models. Nevertheless, the proposed methods needs no target data during training of the models and instead it process target data during the prediction stage which enables the method to be employed in more challenging problems.

In this section, first, formulation of TL problem and assumptions are briefly reviewed. Afterward, our proposed framework for TL is elaborated. 

\subsection{Problem formulation}
 The \textit{source domain} is defined as $D_S=\{\chi_S,P(X_S )\}$, where $ \chi_S$ is the feature space, $X_S=\{x_{S1},…,x_{Sn}\}$, $x_{Si}\in \chi_S$ the data and $P(X_S)$ the marginal distribution from which the source samples are drawn. The corresponding ground truth is defined as $Y_S=\{y_{S1},…,y_{Sn}\}$,  $y_{Si}\in Y_S$, where $Y_S$ is the label space. It is assumed that enough labeled data is available in the \textit{source domain} to find a $\hat{f}_S (x)$ that predicts labels, $y_S$, based on $P_S (y|x)$. Actually, $\hat{f}_S (x)$ aims to approximate the optimal function ${f}_S(x)$. 
 
 In a similar way, a \textit{target domain} is defined as $D_T=\{\chi_S,P(X_T )\}$, where $\chi_T = \chi_S$. The assumption is that no labeled data from the \textit{target domain} is available. On the other hand, $\hat{f}_S (.)$ might not be an appropriate function for approximating ${f}_T (x)$, $x\in \chi_T$, since $P(X_S )\ne P(X_T )$. As a result, if the models trained in the\textit{ source domain} be directly applied to the \textit{target domain}, it could not predict target labels accurately. In this regard, there is a need for a TL method that extract knowledge from the\textit{ source domain} and adapt it to the \textit{target domain}.

\subsection{Between-domain instance transition via Gibbs sampling}
\begin{figure}[htbp]
\centerline{\includegraphics[width=3in]{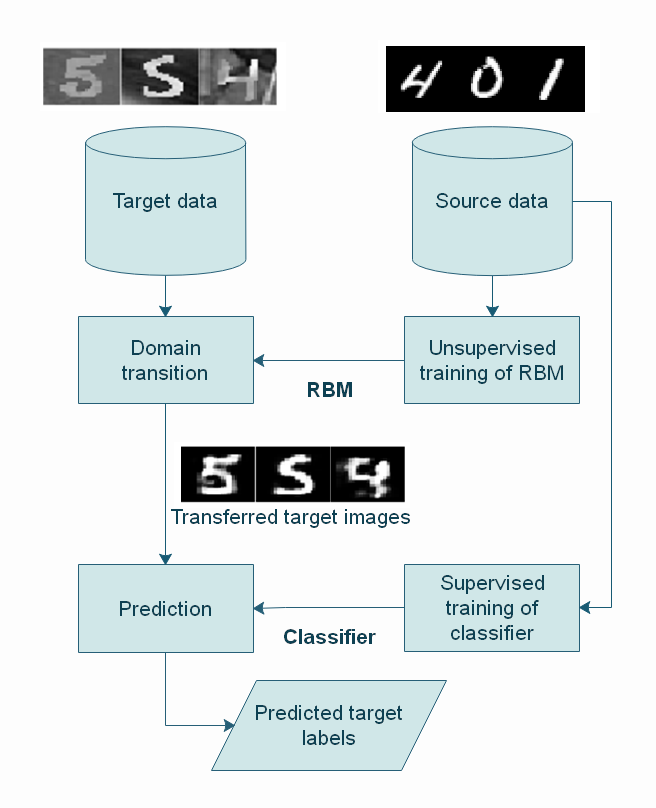}}
\caption{Diagram of the proposed TL framework using Gibbs sampling process in RBM.}
\label{diag}
\end{figure}

The diagrammatic representation of the proposed method is shown in Fig. \ref{diag}. We train a classifier and a RBM using the source data. The classifier predicts labels of given samples based on $P(Y_S|X_S)$.
The RBM which its energy surface is fitted to the source data distribution, by some means understand $P(X_S)$ and is able to extract the concepts behind source samples. Also it is able to generate samples from this distribution. The proposed framework utilize the RBM as a generative model to produce source-like data out of target samples. Next, newly cast samples, which are denoted as $X'_T$, are feed to source classifier for label predictions.

In TL problems, although the probability distribution of source and target data is not equal, $P(X_T) \neq P(X_S)$, target samples are expected to follow a distribution pattern similar to the source one. Therefore, target instances are expected to be in a close neighborhood of source data in terms of euclidean distance. We suggest to think of target data as source instances whose value are altered. Therefore, target instances are assumed to be near local minimums on energy surface of the RBM that is fitted to the source data. As mentioned before, during the process of Gibbs sampling, the current state is likely to transfer to nearby energy minima. Consequently, we deployed Gibbs sampling in RBM in order to transfer target instances to energy minimum points which are corresponding to source samples with the same label. Also, the probability distribution of newly cast samples is close to the probability distribution of source data, $P(X'_T) \simeq P(X_S)$.

As mentioned before, CD algorithm do not contribute to computation of exact gradients of the log-likelihood ,thus, models trained by this algorithm are not complete density models. Since the RBM that we use in our method is better to represents the source data distribution as accurately as possible, it is strongly suggested to use PCD algorithm for training the RBM and avoid using of CD algorithm.

        \section{Results and Discussion}

\begin{figure}[htbp]
\centerline{\includegraphics[width=3.5in]{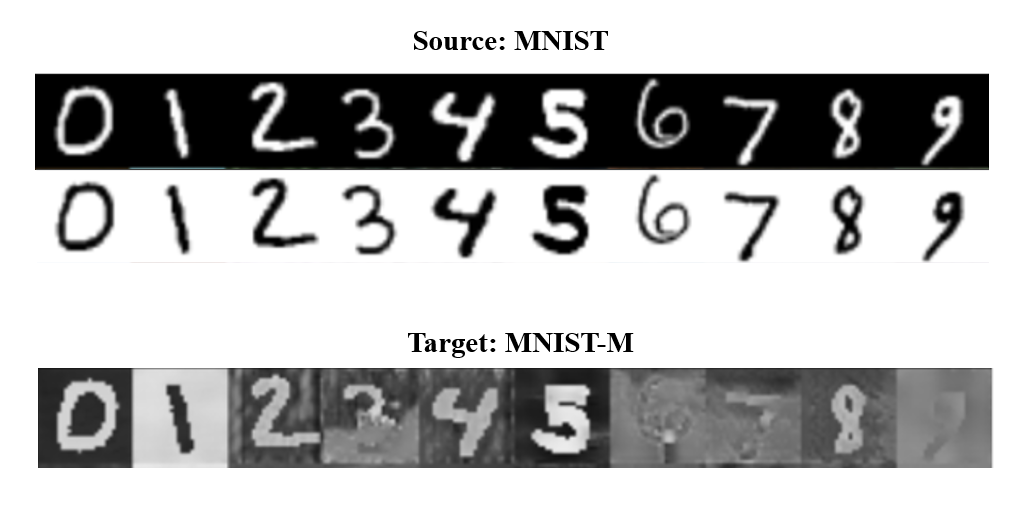}}
\caption{Examples of source and target images.}
\label{sourc-targe}
\end{figure}

In this part, we used the well-known hand-writing datasets that have been used by TL researchers for evaluation of TL algorithms \cite{ganin2016domain, cao2018dida, li2018extracting}. Hand-written digit images \cite{lecun2010mnist}, commonly referred to as MNIST, is the dataset that is used as the source data, $X_S$. MNIST-M , which is used as target data, $X_T$, is a set of data that is made by adding set of random background images to the MNIST images. We used MNIST-M as the target domain. Some instances from these two domains are depicted in Fig. \ref{sourc-targe}.

Originally MNIST samples are images with black background and white digit. We use a gray-scale version of the images in this example. On the other hand, a black number in white background is similar to a white number in a black background. So we concatenate the source data with the inversion of the same data. In other words, we also have handwritten images that their background and digit color are white and black respectively as shown in Fig. \ref{sourc-targe}.

 The size of images is 28*28, which means they consist of 784 pixels. Therefore, the RBM is needed to have 784 units in visible layer. Also, 400 units are used in hidden layer. The RBM is trained by PCD algorithm in the source domain. 

\begin{figure}[htbp]
\centerline{\includegraphics[width=3.5in]{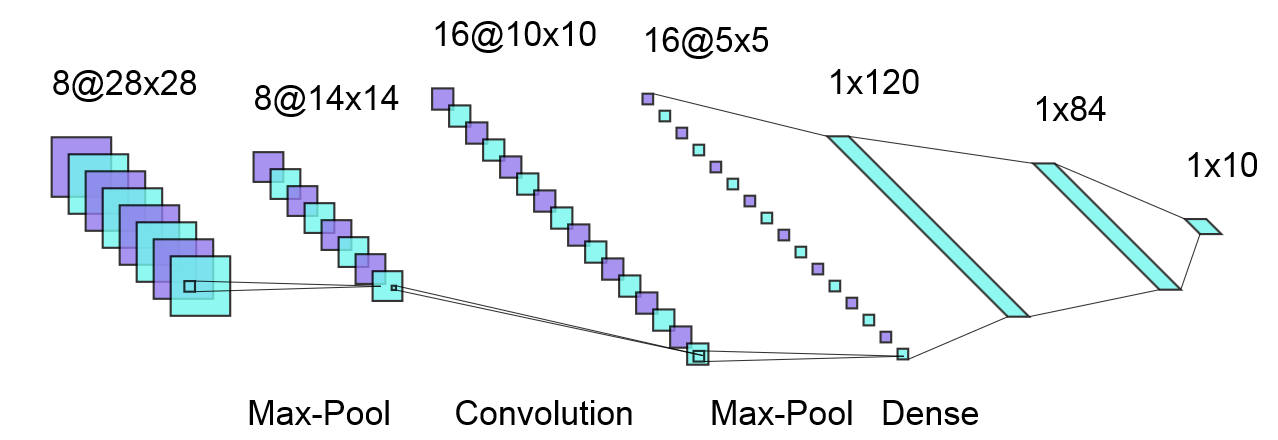}}
\caption{The architecture of the CNN used as classifier.}
\label{cnn}
\end{figure}

A Convolutional Neural Network (CNN) is trained in the source domain in order to predict the label of images. The CNN architecture includes one input layer, one output layer, two convolutional layers, two polling layer and finally two fully connected layers as Fig \ref{cnn}.

The classification accuracy of the model in the source domain is about 98\%, But when the model is applied directly to the MNIST-M target data without TL it can only achieve 38\% accuracy. So, we applied the proposed RBM-based TL. 

\begin{figure}
\centerline{\includegraphics[width=3.65in]{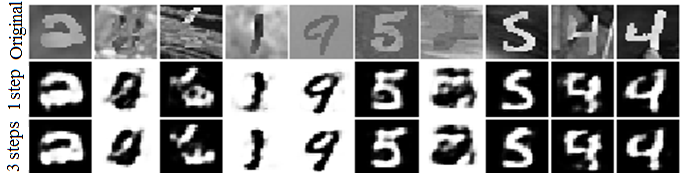}}
\caption{Transferring Target instates to a new domain by 1 and 3 steps Gibbs sampling process with the RBM trained in the source domain.}
\label{after_gs}
\end{figure}

Fig. \ref{after_gs} depicts examples of generated samples,  $X'_T$, as the result of transition of target samples, $X_T$, via 1 and 3 steps of Gibbs sampling using the RBM trained on the source domain. Using TL, classifier trained in the source domain can classify this new images with 65\% accuracy. In many cases, the RBM can make new samples that are similar to source instances with the same labels by transferring target instances to a point near to its minimum energy.
In this experience, it is demonstrated that the proposed method can also perform pixel-level TL since the generated samples are quit similar to MNIST images.

Table \ref{result} provides a comparison between different classifications accuracy to illustrated that how our method have successfully performed TL. In this problem, using our method, the classification accuracy in the MNIST-M increased from 38\% to 65\%. In a case in which a classifier is trained with labeled MNIST-M images, the best achieved classification accuracy is 94\% which shows that classification in the target domain is relatively more challenging compering to the source domain with 98\% accuracy, yet our method can apply the knowledge achieved in the source domain to the target domain.

\begin{table}[]
\begin{center}
\caption{A comparison between classification accuracy in source and target domains under different conditions }\label{result}
\begin{tabular}{|c|l|c|}

\hline
Domain          & \multicolumn{1}{c|}{Training}       & Classification accuracy \\ \hline
Source          & Training the model in source domain & 0.98                    \\
Target          & Using the source model without TL   & 0.38                    \\
\textbf{Target} & \textbf{Using the proposed method with TL}  & \textbf{0.65}           \\
Target          & Training the model in target domain & 0.94                    \\ \hline

\end{tabular}
\end{center}
\end{table}
. 
\section{Conclusion}
Nowadays, researches have found transfer learning a breakthrough in the area of machine learning which is evidenced by massive academic papers being published in this field \cite{weiss2016survey}. While most of the TL methods are based on transferring the information based on the inherent properties of ANN models, especially CNN, this paper proposes a novel method to transfer the information from the source to the target through a restricted Boltzmann machine. 

 The basis of our method is to transfer target instances to source domain by employing Gibbs sampling in an energy based model. The proposed methods is evaluated using well-known MNIST and MNIST-M datasets that are used for TL problems. It successfully improve the classification accuracy in the target domain. 
Besides, the method have the advantage over common domain adaptation methods that it needs no target data during the process of training of models. Actually, the method process target instances and transfers the learned information from source only during the prediction stage. Finally, the method is also employed for pixel-level TL which is quit rare in TL researches.

\bibliographystyle{IEEEtran}
\bibliography{Main.bib}

\end{document}